\documentclass{article}

\usepackage[preprint]{neurips_2026}

\usepackage[utf8]{inputenc} 
\usepackage[T1]{fontenc}    
\usepackage{hyperref}       
\usepackage{url}            
\usepackage{booktabs}       
\usepackage{amsfonts}       
\usepackage{nicefrac}       
\usepackage{microtype}      
\usepackage{xcolor}         
\usepackage{amsmath}
\usepackage{tikz}
\usetikzlibrary{positioning,arrows.meta,shapes.geometric,fit,backgrounds}
\usepackage{soul}
\usepackage{enumitem}
\usepackage{wrapfig}

\title{The CIFAR Synthetic Evidence Corpus for Detecting AI-Generated Evidence}
\author{%
Kelly McConvey$^1$ \quad
Jalehsadat Mahdavimoghaddam$^1$ \quad
Nima Jamali$^2$ \quad
Maksym Taranukhin$^{3,4}$ \quad\\
\textbf {Sajad Ebrahimi}$^1$ \quad
\textbf{Wentao Zhang}$^2$ \quad
\textbf{Karen Eltis}$^5$ \quad 
\textbf{Yuntian Deng}$^2$ \quad\\
\textbf{Maura Grossman}$^2$ \quad 
\textbf{Vered Shwartz}$^{3,4}$ \quad
\textbf{Ebrahim Bagheri}$^1$ \\
$^1$University of Toronto \quad
$^2$University of Waterloo \quad
$^3$University of British Columbia \quad\\
$^4$Vector Institute \quad
$^5$University of Ottawa\\
\texttt{\{s.ebrahimi, kelly.mcconvey, jaleh.mahdavimoghaddam, ebrahim.bagheri\}@utoronto.ca} \\
\texttt{\{nima.jamali, w564zhan, maura.grossman, yuntian\}@uwaterloo.ca}\\
\texttt{\{maksymt, vshwartz\}@cs.ubc.ca}\\
\texttt{karen.eltis@uottawa.ca}\\
}

\begin{document}
\maketitle
\begin{abstract}
The growing ability of generative models to produce realistic documents poses a direct challenge to evidentiary workflows in the justice system and the courts, where decisions increasingly depend on the authenticity of evidence such as receipts, communications, and administrative records. Unlike social media or academic settings, evidentiary documents are often only subtly altered, with small, localized edits that preserve overall plausibility while changing legal meaning. Yet progress on automated detection remains limited, largely due to the absence of suitable training and evaluation data especially suited for the justice system requirements. Existing resources are either focused on photos of human faces or natural scenery or on narrowly scoped academic or social media document types, and do not capture the structure, diversity, or manipulation patterns characteristic of real-world evidentiary data. As a result, current detection systems do not necessarily learn meaningful signals appropriate for the justice system. We introduce the \textit{CIFAR Synthetic Evidence Corpus}, a dataset designed to enable rigorous evaluation of evidence verification under realistic and controlled conditions. The corpus spans multiple document families and a spectrum of manipulation strategies, from small field-level edits to complete document fabrication, and is constructed using a diverse set of state-of-the-art generative tools. It is organized to systematically vary both manipulation complexity and generation method, while enforcing source-level separation between training and test data to reflect real-world generalization challenges. Beyond scale, the dataset is designed to support analysis of model behavior rather than only aggregate accuracy. Each artifact is paired with structured metadata that enables controlled experimentation, including the ability to probe the influence of potential shortcuts and spurious correlations. The corpus also incorporates safeguards to prevent misuse of generated artifacts as real evidence. We accompany the dataset with a benchmark suite evaluating in-domain performance, robustness to distribution shift, and cross-generator generalization. The dataset and evaluation framework provide a foundation for studying the reliability of evidence within the justice system in high-stakes legal contexts.
\end{abstract}
\section{Introduction}
\label{sec:Introduction}

A self-represented litigant with a frontier-model subscription can, in a single afternoon, produce a backdated receipt with internally consistent arithmetic, an email reply impersonating an existing thread participant, or a wholly fabricated business letter on a procedurally generated letterhead. Courts have begun to encounter the consequences. In \emph{Huang v.\ Tesla}, \emph{Valenti v.\ Dfinity}, and \emph{State v.\ Rittenhouse}, U.S.\ courts confronted disputes over the authenticity of audiovisual and documentary evidence without any principled framework for adjudicating them \citep{delfinoDeepfakesTrialCall2023,dalalDeepfakesCourtHow2025}. The exposure is unevenly distributed. Self-represented and under-resourced litigants are simultaneously the population most likely to encounter fabricated submissions and the population least equipped to detect them, with lay accuracy on AI-generated images converging near 62\% and educational interventions providing little reliable improvement \citep{bray_testing_2023,roca_how_2025}. Detection infrastructure is therefore not an optional addition to the courtroom toolkit but a precondition for procedural fairness, and the training data on which such infrastructure depends does not yet exist.

The synthetic-content detection literature has invested heavily in face-deepfake corpora, producing millions of labeled examples across FaceForensics++ \citep{rosslerFaceForensicsLearningDetect2019}, Celeb-DF \citep{liCelebDFLargescaleChallenging2020}, DFDC \citep{dolhansky_deepfake_2020}, and others. Document-forgery resources, by contrast, amount to a few thousand examples concentrated almost entirely on receipts and payslips \citep{tornes_receipt_2023,sidere_dataset_2017,artaud_find_2018}. Beyond their limited scale, existing public corpora suffer from three additional weaknesses that limit rigorous analysis of detector behavior. \emph{First}, they lack coordinated variation across document types, manipulation strategies, and generator families, precluding the controlled comparisons needed to determine whether detection models learn generalizable signals of forgery or merely exploit artifacts tied to specific tools or data sources. \emph{Second}, they largely overlook the localized, single-field edits most likely to arise in legal and evidentiary settings, where the vast majority of a document remains unchanged and only a small region is manipulated, despite real-world tampering typically involving exactly these subtle, targeted alterations designed to preserve overall integrity while evading scrutiny. \emph{Third}, they lack safeguards against shortcut learning and data leakage, allowing models to rely on unintended correlations or spurious cues. The combined effect is that reported performance is difficult to interpret and may overstate true capability, leaving open whether observed accuracy reflects genuine forensic understanding or dependence on incidental patterns in the data.

We introduce the \textbf{CIFAR Synthetic Evidence Corpus}, a corpus of synthetic and authentic documentary evidence built to close these gaps. The corpus spans three documentary-evidence families (receipts, emails, and business and administrative documents), four sophistication tiers ranging from single-field edits to whole-document fabrication, and different generator families covering the multimodal tools that self-represented litigants actually use (GPT, Gemini, Ideogram) together with an open-weights diffusion stack auditable end-to-end. These dimensions yield 1{,}440 manipulated artifacts alongside 1{,}050 clean controls, organized as a fully crossed domain--tier--generator matrix.

The corpus is constructed around three design commitments that prior document-forgery datasets do not jointly satisfy. \emph{First}, training and test pools are partitioned at the source-dataset level rather than the item level, ensuring that any generalization gap at evaluation reflects domain shift rather than item-level memorization. \emph{Second}, every artifact carries a manifest entry recording the generator, the prompt, and the per-item seeds for identity, signature style, and letterhead, enabling evaluators to isolate each potential shortcut's contribution to detector performance through targeted ablation. 
In summary, the paper contributes a 2{,}490-item documentary-evidence corpus structured along a controlled domain--tier--generator matrix that no prior public dataset spans. 

\section{Background and Motivation}\label{sec:Background}

\subsection{Synthetic evidence in legal proceedings}

Courts have already encountered fabricated case citations from large language models, authenticity challenges to exhibits in civil litigation, and growing recourse to what \citet{chesneyDeepFakesLooming2018} named the \emph{liar's dividend}, the bad-faith strategy of denying genuine evidence as fabricated on the basis that fabrication has now become plausible. Recent U.S. cases show how unprepared the courts are for these disputes. In \emph{Huang v.\ Tesla}, the court rebuked a baseless deepfake challenge to authentic video. In \emph{Valenti v.\ Dfinity}, the court dismissed a similar challenge as bad-faith litigation tactics. In \emph{State v.\ Rittenhouse}, the court required expert testimony that a video-zoom function had not altered the underlying footage. None of these decisions rests on a principled framework, and none of the verification tools the field has produced is adequate to the evidentiary, procedural, and interpretive demands courts impose \citep{grimm_artificial_2021,delfinoDeepfakesTrialCall2023,dalalDeepfakesCourtHow2025}.

This risk does not fall on everyone equally. Self-represented and under-resourced litigants, who increasingly rely on freely available LLMs to prepare filings they could not otherwise afford to produce, are both the most likely to inadvertently submit synthetic materials and the least equipped to detect fabrications submitted against them. Research on human detection is not encouraging either where lay accuracy on AI-generated images converges at approximately 62\%, with confidence largely decoupled from accuracy, and educational interventions provide little reliable improvement \citep{bray_testing_2023,roca_how_2025}. Detection tools are therefore no longer optional but a precondition for procedural fairness in the justice system, and the training data on which such tooling depends is the bottleneck that our work in this paper aims to addresses.

\subsection{Why general-purpose detection fails on evidentiary content}

The synthetic-content detection literature has developed primarily in service of two adjacent applications, namely identifying AI-generated text in academic submissions and identifying manipulated media in social-media misinformation. Both settings have shaped available tools in ways that limit their courtroom utility. Three limitations are particularly consequential.

The first concerns the operating environment. Academic and social-media detectors are typically evaluated on inputs that are clean, complete, and presented in standard formats. Evidence reaches courts in a very different state, frequently compressed by the platforms through which it was transmitted, redacted, translated, stripped of metadata during chain-of-custody transfers, or rendered as screenshots of screenshots. \citet{rosslerFaceForensicsLearningDetect2019} showed that detection accuracy on FaceForensics++ collapses from near-perfect at raw quality to near-chance for some manipulation types at differing compression levels, and \citet{chandraDeepfakeEval2024MultiModalIntheWild2025} report that state-of-the-art detectors lose roughly 45 points of AUC when evaluated on real world material relative to academic benchmarks. The degradation from controlled benchmark conditions to real-world deployment is large enough to be determinative of whether a detector is fit for evidentiary use.

The second concerns generational and adversarial fragility. AI generation tools have evolved through distinct technological approaches, first GAN-based and then diffusion-based, each leaving different statistical traces. Detectors trained on one approach generalize within it but fail across. \citet{corviDetectionSyntheticImages2022} confirmed empirically what \citet{wangCNNgeneratedImagesAre2020} had earlier predicted, that GAN-trained detectors transfer across GAN architectures but collapse on diffusion content. \citet{mavali_adversarial_2025} further show that imperceptible perturbations crafted against state-of-the-art image detectors are sufficient to cause misclassification, and the analogous finding holds for AI-text detectors \citep{huang_are_2024}. A detector validated against today's generators on benign inputs offers no guarantee of performance against future generators or against motivated evasion.

The third concerns the form of the output. Most current detection systems return a binary label, a confidence score, or a saliency heatmap. None of these maps onto evidentiary reasoning, which works in terms of admissibility, weight, burden allocation, calibrated abstention, and the structured justifications those determinations require \citep{dalalDeepfakesCourtHow2025,grossman_judicial_2025}. Producing detectors with courtroom-compatible outputs is not solely an architectural problem but also more deeply a dataset design problem. Training data must record manipulation types, generator identity, and ground-truth metadata at a granularity that supports more than a binary label, and existing benchmarks do not provide that.

\subsection{Existing datasets and their limitations}


\textbf{Forgery-labeled benchmarks.} Face-deepfake corpora are abundant. FaceForensics++ \citep{rosslerFaceForensicsLearningDetect2019}, Celeb-DF \citep{liCelebDFLargescaleChallenging2020}, the DeepFake Detection Challenge \citep{dolhansky_deepfake_2020}, ForgeryNet \citep{he_forgerynet_2021}, and OpenForensics \citep{le_openforensics_2021} together provide millions of labeled face-manipulation examples. General-scene image-tampering benchmarks include CASIA \citep{dong_casia_2013} and NIST OpenMFC \citep{guan_mfc_2019}. Document-forgery resources are by contrast sparse and concentrated on financial documents. The Find It Again! receipt corpus \citep{tornes_receipt_2023} and the earlier ICPR Find It! contest dataset \citep{artaud_find_2018} together provide fewer than 3{,}000 receipt examples with forgery annotations, and the \citet{sidere_dataset_2017} corpus contributes 477 payslips. The total volume of document-forgery training data across all publicly available sources is on the order of a few thousand examples, against several million for face deepfakes, an imbalance that does not reflect the distribution of evidentiary submissions in routine litigation but the historical distribution of academic interest.

\textbf{Authentic source corpora.} Detection models for documentary evidence must be built on a foundation of unmanipulated material. SROIE \citep{huang_icdar2019_2019} and CORD \citep{park_cord_2019} provide receipt corpora, RVL-CDIP \citep{harley_evaluation_2015} and DUDE \citep{fink_icdar_2023} cover business and administrative documents, the Enron corpus and the LDC's Avocado collection cover corporate email, and the UCSF Industry Documents Library provides litigation-released business records at scale. None of these carries manipulation labels, so they are usable only as starting material for forgeries that researchers must construct themselves. 

\begin{table}[t]
\centering
\small
\setlength{\tabcolsep}{4pt}
\caption{Comparison of forgery-labeled detection benchmarks discussed in
Section~\ref{sec:Background}.}
\label{tab:dataset_comparison}
\begin{tabular}{llrcccc}
\toprule
Dataset & Domain & Size & Matrix & Local. & Leak. & Prov. \\
\midrule
FaceForensics++ \citep{rosslerFaceForensicsLearningDetect2019}      & Faces      & 1.8M frames   & $\times$ & $\times$    & $\times$ & $\times$ \\
Celeb-DF \citep{liCelebDFLargescaleChallenging2020}                  & Faces      & 5.6K videos   & $\times$ & $\times$    & $\times$ & $\times$ \\
DFDC \citep{dolhansky_deepfake_2020}                                 & Faces      & 100K+ videos  & $\times$ & $\times$    & $\times$ & $\times$ \\
ForgeryNet \citep{he_forgerynet_2021}                                & Faces      & 2.9M images   & $\times$ & $\times$    & $\times$ & $\times$ \\
OpenForensics \citep{le_openforensics_2021}                          & Faces      & 115K images   & $\times$ & $\times$    & $\times$ & $\times$ \\
CASIA \citep{dong_casia_2013}                                        & Scenes     & 12K images    & $\times$ & $\times$    & $\times$ & $\times$ \\
NIST OpenMFC \citep{guan_mfc_2019}                                   & Mixed      & Varies        & $\times$ & $\times$    & $\times$ & $\times$ \\
Find It Again! \citep{tornes_receipt_2023}                           & Receipts   & 163 forgeries & $\times$ & \checkmark  & $\times$ & $\times$ \\
ICPR Find It! \citep{artaud_find_2018}                               & Receipts   & $<$3K total   & $\times$ & \checkmark  & $\times$ & $\times$ \\
\citet{sidere_dataset_2017}                                          & Payslips   & 477 docs      & $\times$ & \checkmark  & $\times$ & $\times$ \\
\midrule
\textbf{CIFAR Synthetic Evidence Corpus (ours)}                      & Documents  & 2{,}490 items & \checkmark & \checkmark & \checkmark & \checkmark \\
\bottomrule
\end{tabular}
\end{table}

\textbf{Gaps.} Several evidentiary categories that are routine in litigation have no public benchmark whatsoever, including doorbell and consumer-camera footage, messaging-platform screenshots (SMS, WhatsApp, social platforms), and identity and legal documents (licenses, contracts, affidavits). Equally consequential is the absence of benchmarks for \emph{localized} manipulations to otherwise authentic documents, the manipulation class most likely to be consequential in court precisely because most of the artifact remains intact, and the class on which human detection performs worst \citep{roca_how_2025}. And no public dataset spans a controlled matrix of document domain, manipulation sophistication, and generator family, the structure required to evaluate whether a detector has learned genuine manipulation traces or merely tool-family fingerprints, and to support the kind of ablation studies that would let a court understand what a detector's performance number actually warrants. Table~\ref{tab:dataset_comparison} summarizes how existing forgery-labeled benchmarks compare on the four design axes the Synthetic Evidence Corpus targets.

\section{Dataset Collection and Generation}\label{sec:dataset}

We construct the Synthetic Evidence Corpus through a two-stage pipeline designed to mirror how a self-represented litigant or unsophisticated defendant might plausibly manipulate documentary evidence using widely available generative AI tools. We first assemble a base set of real documents from established research corpora and public-record collections (Section~\ref{sec:collection}) and then apply a suite of manipulations across four escalating sophistication tiers using four distinct generator families (Section~\ref{sec:generation}).

\subsection{Scope and Inclusion Criteria}\label{sec:scope}

The CIFAR Synthetic Evidence Corpus is deliberately focused along several dimensions. We record those choices explicitly here, both to make the corpus's claims legible and to clarify what the corpus does not attempt to measure.

\textbf{Document artifacts only.} The corpus comprises PDFs and raster images (JPG/PNG) of documents. We exclude photographs of physical objects or scenes, audio recordings, video, and screenshots of text-message or chat interfaces. This scoping reflects the most common forms of documentary evidence encountered in civil and administrative proceedings, namely receipts, correspondence, and business records \citep{grimm_artificial_2021}, and allows the corpus to be internally comparable across items. Extension to other evidentiary modalities is left to future releases.

\textbf{Three document families.} Within the document category, we cover receipts, emails, and business and administrative documents (letters, memos, reports, contracts, and forms). We focus on these three families because they span the core categories of documentary evidence a self-represented litigant is both likely to submit and technically able to manipulate with commodity generative tools, including transactional records (receipts), communications (emails), and formal institutional documents (letters, memos, contracts). This scoping follows from the broader literature on AI-generated evidence in court \citep{grossman_judicial_2025,grimm_artificial_2021}, which documents the proliferation of fabricated documentary artifacts across exactly these categories without enumerating them as a typology. We adopt the enumeration here to delimit the corpus's scope. Specialized genres such as court filings, medical records, and financial statements are excluded from this release because each is evidentiary but each would also require domain-specific sourcing and review arrangements, which is impossible to obtain for public release due to privacy restrictions of such material.


\textbf{Synthetic manipulations only.} With the sole exception of the 163 pre-labeled forgeries from Find It Again! \citep{tornes_receipt_2023}, retained as \texttt{tier=external\_labeled} calibration material, every manipulated item in the Synthetic Evidence Corpus is produced by the generation pipeline described below. The corpus does not contain genuinely forged real-world documents. This choice is driven both by the limited availability of labeled forgery data at scale and by the ethical and legal difficulties of collecting real fraudulent artifacts. We discuss the implications for external validity in Section~\ref{sec:limitations}.

\subsection{Real Dataset Collection}\label{sec:collection}

We draw real documents from eleven publicly available source datasets, organized by document family and pool assignment. Table~\ref{tab:sources} summarizes the source-to-pool mapping. The three document families covered are receipts (RCT), emails (EML), and business and administrative documents (DOC). For each family, source datasets are partitioned into a \emph{training pool} built on research datasets with permissive manipulation licenses, and a \emph{test pool} built on disjoint public-record collections drawn from different underlying populations. This separation is enforced at the source-dataset level rather than only at the item level. Training and test pools draw from non-overlapping corpora, so any generalization gap observed during evaluation reflects domain shift rather than item-level memorization.

\textbf{Receipts} are drawn from SROIE (ICDAR 2019 Task 3) \citep{huang_icdar2019_2019} and CORD-v2 \citep{park_cord_2019}, with a held-back portion of each reserved for the test pool and disjoint document IDs enforced between pools. Find It Again! \citep{tornes_receipt_2023} contributes additional genuine receipts and supplies 163 pre-labeled forgeries that we retain in the manifest as \texttt{tier=external\_labeled} calibration material.

\textbf{Emails} in the training pool are drawn from the Enron Email Corpus (CMU release) and the email class of RVL-CDIP \citep{harley_evaluation_2015}. The Enron corpus is confined strictly to training, and the test pool draws exclusively from the Avocado Research Email Collection (LDC2015T03), which is disjoint from both Enron and RVL-CDIP in its custodian population.

\textbf{Business and administrative documents} in the training pool come from the non-email classes of RVL-CDIP and from DUDE \citep{fink_icdar_2023}. The test pool draws from the letter, memo, and report subsets of the UCSF Industry Documents Library, sourced from the tobacco, drug, food, fossil fuel, and opioid litigation collections.

For each base item we record the source file, its SHA-256, the source dataset and license, the pool assignment, and a \texttt{tier=0} (clean) label. Pilot corpus sizes are 250 clean items per family in the training pool and 100 clean items per family in the test pool. Clean controls are not strictly unmodified. Approximately 10\% of clean items in each family are re-saved through the same image and PDF editors used on the manipulated side without any content edit. This places format-level fingerprints (file structure, compression artifacts, EXIF metadata) on both sides of the label, so that a detector cannot discriminate manipulated from clean items on the basis of tooling artifacts alone.

\begin{table}[t]
\centering
\small
\caption{Source-to-pool mapping. Training and test pools are disjoint at the source-dataset level, not merely at the item level.}
\label{tab:sources}
\begin{tabular}{lll}
\toprule
Family & Training pool & Test pool \\
\midrule
Receipts (RCT) & SROIE, CORD-v2, Find It Again! & SROIE (held-back), CORD-v2 (held-back) \\
Emails (EML) & Enron, RVL-CDIP (email class) & Avocado \\
Business docs (DOC) & RVL-CDIP (non-email), DUDE & UCSF Industry Documents Library \\
\bottomrule
\end{tabular}
\end{table}

\begin{figure}[t]
    \centering
    \includegraphics[width=1\linewidth]{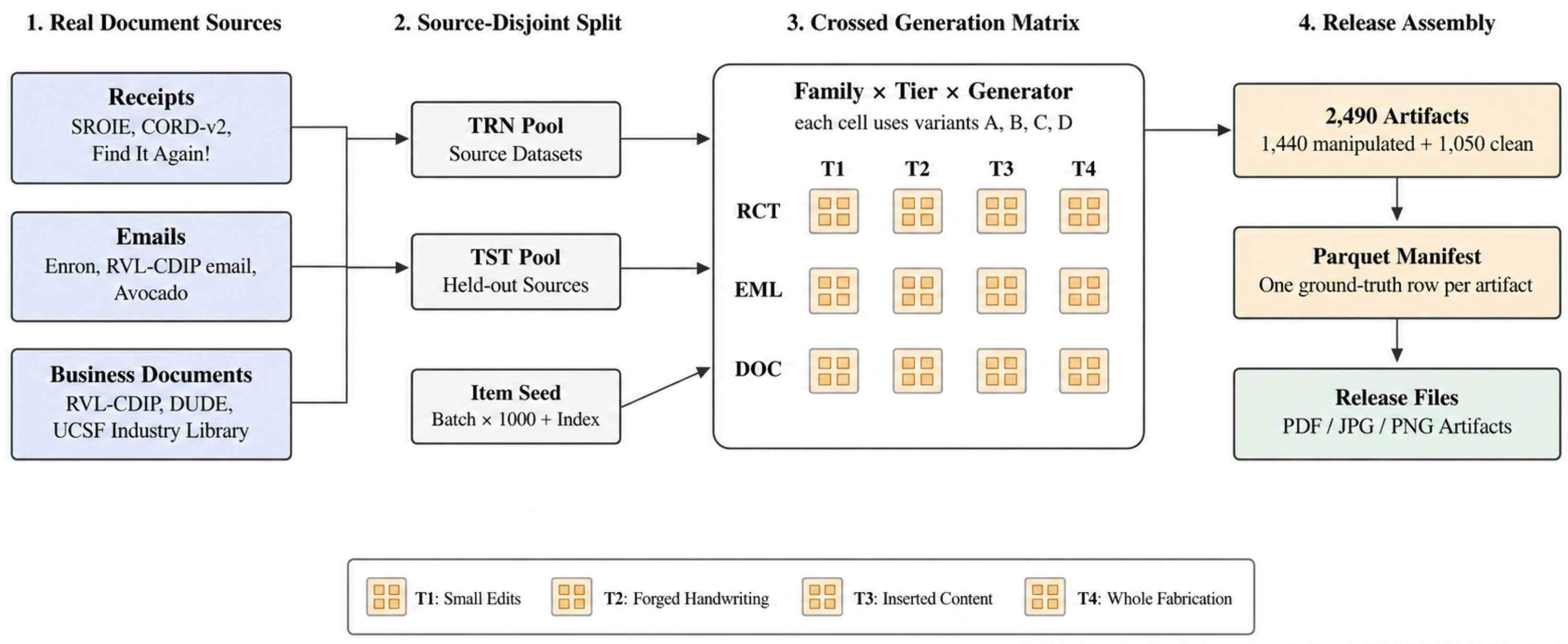}
    \caption{The overview of the process adopted for curating the CIFAR Synthetic Evidence Corpus.}
    \label{fig:generation_process}
\end{figure}
\subsection{Synthetic Dataset Generation}\label{sec:generation}

We generate manipulated artifacts along two orthogonal axes. A \emph{sophistication tier} describes what kind of edit is made, and a \emph{tool variant} describes which generator family produces the edit. The full corpus spans $2 \text{ pools} \times 3 \text{ families} \times 4 \text{ tiers} \times 4 \text{ variants} = 96$ batches, with 20 manipulated items per training batch and 10 per test batch. This yields 960 manipulated training items and 480 manipulated test items, alongside the clean versions described in Section~\ref{sec:collection}. The full pipeline of our work is summarized in Figure~\ref{fig:generation_process}. 

\subsubsection{Sophistication Tiers}

Manipulations are organized into four tiers of increasing effort, chosen to span the realistic range of what an unsophisticated actor with a frontier-model subscription can produce in a few hours. A single manipulated item receives exactly one tier of edit. Stacked manipulations (e.g., a date change combined with a forged signature on the same document) are deferred to a future release.

\textbf{Tier 1 (Small edits)} changes a single field on an otherwise authentic document. The canonical edits are date changes (sampled uniformly from $\{\pm 7, \pm 30, \pm 90, \pm 365, \pm 730\}$ days for receipts and documents, and $\{\pm 1\text{h}, \pm 6\text{h}, \pm 1\text{d}, \pm 7\text{d}, \pm 30\text{d}\}$ for email timestamps), dollar-amount changes (a sampled line item is multiplied by a factor in $\{0.5, 1.5, 2.0, 3.0, 5.0\}$), and timeline manipulations (reordering, deleting, or replacing pages or messages). For dollar edits we include two deliberate sub-variants in equal proportion. The \emph{consistent} sub-variant recomputes subtotals, taxes, and totals, while the \emph{inconsistent} sub-variant updates only the changed line item and the total, leaving the arithmetic internally wrong. The inconsistent sub-variant models the amateur mistake pattern that the deployed detector should learn to exploit.

\textbf{Tier 2 (Forged handwriting)} introduces a signature or handwritten annotation. For each item, a style index is drawn from a pool of 200 signature and handwriting styles, split disjointly into 150 training styles and 50 test styles, with no style appearing in both pools. The chosen style is perturbed per item with a random affine transform (rotation $\pm 2^\circ$, scale $\pm 5\%$, shear $\pm 2^\circ$), ink-color jitter in HSV space, and pen-pressure noise, all sampled deterministically from the item seed. The name associated with a signature is drawn freshly via Faker and is independent of the style index, so that name identity does not leak style identity.

\textbf{Tier 3 (Inserted content)} adds a semantically meaningful piece of new content to an existing document, including a contract clause, a paragraph in a memo or report, a clinical note on a medical-style document, or a new reply in an existing email thread. The inserted content is drafted by a large language model using a fixed prompt template (logged verbatim in the manifest) and is inserted at a plausible location using a PDF or image editor, with font and indentation matched to the surrounding text. The email-insertion case warrants particular mention as the inserted reply impersonates a participant already present in the thread rather than introducing a fresh fictional identity. Dropping a stranger into a real thread produces an obviously wrong training signal, while impersonating an existing participant is both the more realistic case and the form of fabrication a self-represented litigant would actually attempt.

\textbf{Tier 4 (Whole fabrication)} generates a complete document from scratch with no source artifact. Three to five documents from the corresponding source dataset are passed to the generator as style anchors so that the output belongs plausibly to the target genre. Fresh fictional identities (names, employers, addresses, email addresses, phone numbers, and, for medical-style documents, NPIs with explicitly invalid Luhn checksums) are supplied per item via a Faker-based identity library seeded deterministically with $\text{batch\_seed} \times 1000 + \text{item\_index}$. Letterheads for fabricated business documents are produced procedurally per item from 10 layout templates and a pool of 20 system fonts, with no fixed letterhead PNGs reused across items. For fabricated receipts, we include arithmetically consistent and arithmetically broken sub-variants in equal proportions.

\subsubsection{Tool Variants}

Each (pool, family, tier) cell is generated four times, once with each of four generator families. Variants A, B, and C are frontier closed-model multimodal APIs, namely GPT-4o with gpt-image-1 (A), Gemini 2.5 Pro with Gemini 2.5 Flash Image (B), and Ideogram 3.0 (C). These three are selected because they are what self-represented litigants actually use in practice. Variant D is an open-weights diffusion stack (Flux.1-Kontext-dev for localized edits in Tiers 1--3, and Flux.1-dev for whole-image fabrication in Tier 4). Variant D serves two purposes. It extends detector evaluation across the closed-LLM and open-diffusion models, and it is the only variant whose pipeline can be audited end-to-end without reliance on a commercial provider. Exact model versions are pinned in a \texttt{tools.yaml} file referenced from every manifest entry, so that any future changes in the underlying APIs is recoverable from the corpus version.


\subsubsection{Shortcut-leakage Controls}

A central design concern is that any feature systematically present on the manipulated side but absent from the clean side risks becoming a shortcut that a detector learns in lieu of the actual forgery signal. We apply four controls throughout the generation pipeline to reduce this risk:

\begin{itemize}[label=--, leftmargin=0.5cm]
\item \emph{Per-item identities.} Every fictional identity used anywhere in the corpus, including names, employers, addresses, email addresses, phone numbers, and account numbers, is generated freshly per item via a Faker-based library seeded with $\text{batch\_seed} \times 1000 + \text{item\_index}$. No identity recurs across items, even within the same batch. This preserves full reproducibility (a given seed always produces the same identity) while preventing the detector from memorizing recurring personas.

\item \emph{Disjoint style pools for Tier 2.} As noted above, the 200-style signature and handwriting pool is partitioned into 150 training styles and 50 test styles with no overlap. Within each pool, the assigned style for an item is perturbed per item to prevent pixel-level style memorization.

\item \emph{Parametric letterheads.} Letterheads for Tier 4 business documents are generated procedurally from per-item seeds rather than drawn from a fixed set of PNGs. Template, font, color scheme, and logo geometry are all sampled deterministically from the item seed.

\item \emph{Balanced format fingerprints.} Approximately 10\% of clean controls per family are re-saved through the same image and PDF editors used on the manipulated side, without content edits, placing format-level fingerprints on both sides of the label.
\end{itemize}

These controls are constructed to make ablation straightforward and reproducible. For each instance, the manifest stores the \texttt{identity seed}, \texttt{style pool}, and \texttt{letterhead seed}, which allows the corpus to be regenerated with targeted substitutions such as identities, letterheads, fonts, or style pools while keeping other factors fixed. This enables direct attribution of performance differences to specific factors and provides a clear test of whether detectors rely on genuine forgery signals or incidental cues.

\subsubsection{Provenance Marking}

Every artifact in the corpus, clean or manipulated, carries an embedded marker identifying it as Synthetic Evidence Corpus material. The marker takes three redundant forms, namely an EXIF/XMP tag with key \texttt{synthetic-evidence-corpus} and value \texttt{true} for images that retain metadata, a steganographic flag for images whose EXIF would be stripped by downstream processing, and a sentinel hash logged in the manifest. At least one of the three markers is designed to survive a re-save through a consumer image editor. The intent is to make it difficult for a corpus artifact to be mistaken for genuine evidence if it escapes the research environment, while not interfering with detector training. Markers are applied identically to both clean and manipulated items.

\section{Dataset Description}\label{sec:data_stats}
\begin{wrapfigure}{r}{0.5\textwidth}
    \centering
    \vspace{-\baselineskip}
    \includegraphics[width=0.48\textwidth]{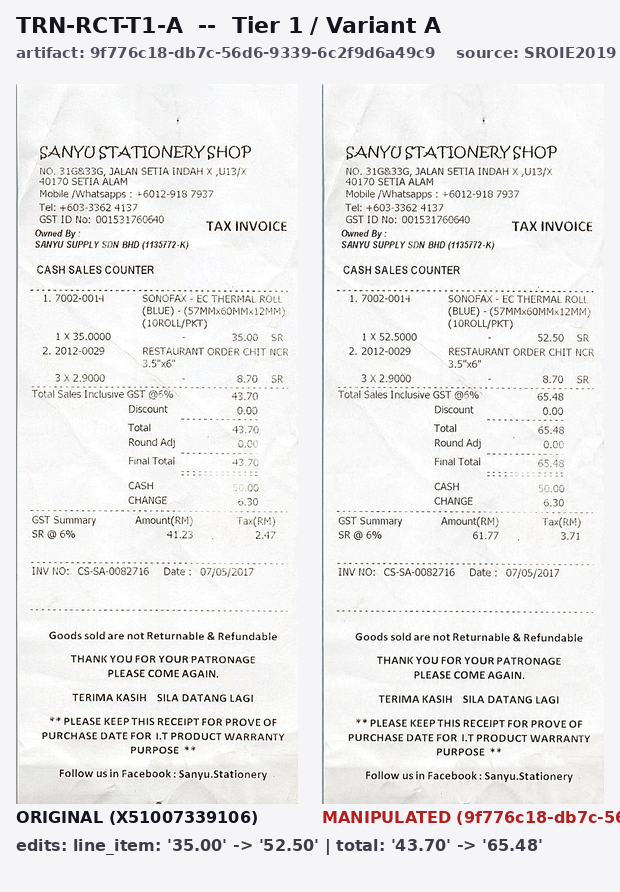}
    \caption{Example Tier 1 manipulation. A SROIE receipt with a single-field edit produced by Variant A.}
    \label{fig:example}
\end{wrapfigure}

The CIFAR Synthetic Evidence Corpus comprises 2{,}490 documentary artifacts, i.e., 1{,}050 clean controls and 1{,}440 manipulated items. Table~\ref{tab:composition} reports the breakdown by pool, family, tier, and generator variant. Clean controls are distributed evenly across families (250 per family in the training pool, 100 per family in the test pool), with approximately 10\% in each family re-saved  through the same image and PDF editors used on the manipulated side to balance format-level fingerprints across the label. Manipulated items are produced through the fully crossed $\text{family} \times \text{tier} \times \text{variant}$ matrix described in Section~\ref{sec:generation}, with 20 items per cell in the training pool and 10 items per cell in the test pool, yielding 960 manipulated training items and 480 manipulated test items. The 163 pre-labeled forgeries from Find It Again! \citep{tornes_receipt_2023}, retained as \texttt{tier=external\_labeled} calibration material, are reported separately and are not part of the manipulated counts above.

Each artifact is paired with a row in a single Parquet manifest recording the source artifact identifier (null for Tier~4), source dataset and license, generator family and pinned model version, the full prompt where applicable, pixel- or character-level edit regions, the per-item identity, style-pool, and letterhead seeds, a free-text description of the intended evidentiary role, the provenance-marker confirmation, and a SHA-256 of the final file. The same seeds that deterministically produce each artifact also support the ablation studies described in Section~\ref{sec:generation}. The corpus is hosted on Harvard Dataverse at \url{https://doi.org/10.7910/DVN/YY0IUH}. Generation code is available at \url{https://github.com/UofT-CIFAR/Synthetic-Evidence-Document-Corpus}.

\begin{table}[t]
\centering
\small
\setlength{\tabcolsep}{5pt}
\caption{Composition of the CIFAR Synthetic Evidence Corpus by pool, family, tier, and generator variant. Each manipulated cell is generated four times, once per variant (A: GPT-4o + gpt-image-1, B:
Gemini 2.5, C: Ideogram 3.0; D: Flux.1-Kontext-dev / Flux.1-dev). Training cells contain 20 items per variant, test cells contain 10. Clean controls (T0) are not generated per variant. The 163 external pre-labeled forgeries from Find It Again! are reported separately as calibration material.}
\label{tab:composition}
\begin{tabular}{llrrrrrr}
\toprule
& & \multicolumn{4}{c}{Manipulated (per variant)} & & \\
\cmidrule(lr){3-6}
Pool & Family & T1 & T2 & T3 & T4 & Clean (T0) & Total \\
\midrule
Training (TRN) & Receipts (RCT)      & 20 & 20 & 20 & 20 & 250 & 570 \\
               & Emails (EML)        & 20 & 20 & 20 & 20 & 250 & 570 \\
               & Business docs (DOC) & 20 & 20 & 20 & 20 & 250 & 570 \\
\midrule
Test (TST)     & Receipts (RCT)      & 10 & 10 & 10 & 10 & 100 & 240 \\
               & Emails (EML)        & 10 & 10 & 10 & 10 & 100 & 240 \\
               & Business docs (DOC) & 10 & 10 & 10 & 10 & 100 & 240 \\
\midrule
\multicolumn{2}{l}{\textbf{Subtotal}} & 360 & 360 & 360 & 360 & 1{,}050 & 2{,}490 \\
\midrule
\multicolumn{2}{l}{External calibration (Find It Again!)} & \multicolumn{4}{c}{163 pre-labeled forgeries} & --- & 163 \\
\bottomrule
\end{tabular}
\end{table}

\section{Discussion, Limitations and Future Work}\label{sec:discussion} \label{sec:limitations}

\textbf{Scope limitations.} The corpus covers three document families, and a single manipulation per item. Court filings, medical records, text messages and financial statements, each evidentiary in routine litigation, are absent, as are stacked manipulations of the kind a sophisticated actor might combine on a single document (a date change, a forged signature, and an inserted clause appearing together). The corpus covers documentary evidence only.

\textbf{Synthetic rather than real-world fraud.} With the exception of the 163 pre-labelled forgeries from Find It Again!
\citep{tornes_receipt_2023}, every manipulated item in our corpus is produced by the generation pipeline of section~\ref{sec:generation}. The corpus therefore measures detector performance against the kinds of manipulations a self-represented litigant with commodity tools could plausibly produce, not against the population of forgeries actually encountered in court. Real evidentiary fraud may differ from what we have captured.

\textbf{Tool-version drift.} Variants A--C are commercial APIs whose underlying models change without notice. We pin exact model versions in \texttt{tools.yaml} and record the version used per item in the manifest, so that any drift is at minimum traceable, but artifacts generated against a 2025-pinned GPT-4o are not equivalent to artifacts a researcher would obtain from the same API endpoint a year later. Variant D (Flux.1-Kontext-dev / Flux.1-dev) is the only branch of the corpus fully reproducible without commercial-providers.

\textbf{Lab-to-wild gap.} The lab-to-wild performance gap documented for face-deepfake detectors (on the order of 45 AUC points between academic benchmarks and real social-media content \citep{chandraDeepfakeEval2024MultiModalIntheWild2025}) has not been quantified for documentary-evidence detection, but there is no principled reason to expect it to be smaller. Substrate selection (Section~\ref{sec:collection}) mitigates this by drawing test-pool material from corpora disjoint from training-pool sources, but documents reaching courts have typically been compressed, redacted, scanned, and re-filed in ways that no public corpus fully reproduces. Performance on the Synthetic Evidence Corpus's test pool is an upper bound on performance under realistic deployment conditions.

\textbf{Single-artifact analysis.} Real evidentiary submissions are often interlinked packages whose internal cross-document consistency is part of what must be verified (a contract whose clauses contradict the correspondence that introduces it, an affidavit whose timeline is inconsistent with attached exhibits). The corpus does not support evaluation of cross-document consistency checks.

\textbf{Future Work.} \label{sec:future_work}
The most direct extensions of the CIFAR Synthetic Evidence Corpus address the scope limitations above. (1) \emph{Stacked manipulations}, in which two or more edits are applied to a single document with manifest entries recording each edit independently, would let detectors learn to flag combinations that no individual tier captures. (2) \emph{Cross-modal packages} of mutually referencing documents (a contract paired with an email thread discussing it, an affidavit paired with photographs whose dates and locations are recorded in its text) would extend the corpus toward the consistency-checking task that real evidentiary review involves, building on cross-modal directions sketched in recent work on vision-language alignment for forgery detection. (3) \emph{Additional document families} such as court filings, medical records, and financial statements would broaden coverage, though each requires domain-specific substrate sourcing and review arrangements that we defer to future releases.

\section{Concluding Remarks}\label{sec:Conclusion}

The CIFAR Synthetic Evidence Corpus is a step toward detection infrastructure that the justice system can actually use. By covering three documentary-evidence families across four manipulation tiers and four generator families, the corpus provides the structured variation that prior document-forgery datasets do not offer, and that meaningful evaluation of detector reliability requires. We hope the corpus, its manifest, and its provenance markers provide a foundation that other researchers can extend, and build upon.

\bibliographystyle{plainnat}
\bibliography{references}

@inproceedings{park_cord_2019,
	address = {Vancouver, Canada},
	title = {{CORD}: {A} {Consolidated} {Receipt} {Dataset} for {Post}-{OCR} {Parsing}},
	url = {{https://openreview.net/forum?id=SJl3z659UH}},
	language = {en},
	booktitle = {Workshop on {Document} {Intelligence} at {NeurIPS} 2019},
	author = {Park, Seunghyun and Shin, Seung and Lee, Bado and Lee, Junyeop and Surh, Jaeheung and Seo, Minjoon and Lee, Hwalsuk},
	year = {2019},
}

@misc{fink_icdar_2023,
	address = {Cham},
	title = {{ICDAR} 2023 {Competition} on {Document} {UnderstanDing} of {Everything} ({DUDE})},
	url = {https://link.springer.com/10.1007/978-3-031-41679-8_24},
	doi = {10.1007/978-3-031-41679-8_24},
	language = {en},
	urldate = {2026-02-26},
	publisher = {Springer Nature Switzerland},
	author = {Van Landeghem, Jordy and Tito, Rubèn and Borchmann, Łukasz and Pietruszka, Michał and Jurkiewicz, Dawid and Powalski, Rafał and Józiak, Paweł and Biswas, Sanket and Coustaty, Mickaël and Stanisławek, Tomasz},
	collaborator = {Fink, Gernot A. and Jain, Rajiv and Kise, Koichi and Zanibbi, Richard},
	year = {2023},
}

@misc{mavali_adversarial_2025,
	title = {Adversarial {Robustness} of {AI}-{Generated} {Image} {Detectors} in the {Real} {World}},
	url = {http://arxiv.org/abs/2410.01574},
	doi = {10.48550/arXiv.2410.01574},
	urldate = {2026-04-30},
	publisher = {arXiv},
	author = {Mavali, Sina and Ricker, Jonas and Pape, David and Fischer, Asja and Schönherr, Lea},
	month = jun,
	year = {2025},
	note = {arXiv:2410.01574 [cs]},
}

@misc{huang_are_2024,
	title = {Are {AI}-{Generated} {Text} {Detectors} {Robust} to {Adversarial} {Perturbations}?},
	url = {http://arxiv.org/abs/2406.01179},
	doi = {10.48550/arXiv.2406.01179},
	urldate = {2026-04-30},
	publisher = {arXiv},
	author = {Huang, Guanhua and Zhang, Yuchen and Li, Zhe and You, Yongjian and Wang, Mingze and Yang, Zhouwang},
	month = jun,
	year = {2024},
	note = {arXiv:2406.01179 [cs]},
}

@article{grossman_judicial_2025,
	title = {Judicial {Approaches} to {Acknowledged} and {Unacknowledged} {AI}-{Generated} {Evidence}},
	volume = {26},
	copyright = {Copyright (c) 2025 Maura Grossman, Paul Grimm},
	issn = {1938-0976},
	url = {https://journals.library.columbia.edu/index.php/stlr/article/view/13890},
	doi = {10.52214/stlr.v26i2.13890},
	language = {en},
	number = {2},
	urldate = {2026-04-23},
	journal = {Science and Technology Law Review},
	author = {Grossman, Maura and Grimm, Paul},
	month = may,
	year = {2025},
}

@article{grimm_artificial_2021,
	title = {Artificial {Intelligence} as {Evidence}},
	volume = {19},
	issn = {1549-8271},
	url = {https://scholarlycommons.law.northwestern.edu/njtip/vol19/iss1/2},
	number = {1},
	journal = {Northwestern Journal of Technology and Intellectual Property},
	author = {Grimm, Paul and Grossman, Maura and Cormack, Gordon},
	month = dec,
	year = {2021},
	pages = {9},
}

@inproceedings{huang_icdar2019_2019,
	title = {{ICDAR2019} {Competition} on {Scanned} {Receipt} {OCR} and {Information} {Extraction}},
	url = {http://arxiv.org/abs/2103.10213},
	doi = {10.1109/ICDAR.2019.00244},
	urldate = {2026-04-23},
	booktitle = {2019 {International} {Conference} on {Document} {Analysis} and {Recognition} ({ICDAR})},
	author = {Huang, Zheng and Chen, Kai and He, Jianhua and Bai, Xiang and Karatzas, Dimosthenis and Lu, Shjian and Jawahar, C. V.},
	month = sep,
	year = {2019},
	note = {arXiv:2103.10213 [cs]},
	pages = {1516--1520},
}

@misc{harley_evaluation_2015,
	title = {Evaluation of {Deep} {Convolutional} {Nets} for {Document} {Image} {Classification} and {Retrieval}},
	url = {http://arxiv.org/abs/1502.07058},
	doi = {10.48550/arXiv.1502.07058},
	urldate = {2026-02-27},
	publisher = {arXiv},
	author = {Harley, Adam W. and Ufkes, Alex and Derpanis, Konstantinos G.},
	month = feb,
	year = {2015},
	note = {arXiv:1502.07058 [cs]},
}

@inproceedings{artaud_find_2018,
	address = {Beijing, China},
	title = {Find it! {Fraud} {Detection} {Contest} {Report}},
	issn = {1051-4651},
	url = {https://ieeexplore.ieee.org/document/8545428},
	doi = {10.1109/ICPR.2018.8545428},
	urldate = {2026-02-26},
	booktitle = {2018 24th {International} {Conference} on {Pattern} {Recognition} ({ICPR})},
	author = {Artaud, Chloé and Sidère, Nicolas and Doucet, Antoine and Ogier, Jean-Marc and Yooz, Vincent Poulain D'Andecy},
	month = aug,
	year = {2018},
	note = {ISSN: 1051-4651},
	pages = {13--18},
}

@article{guan_mfc_2019,
	title = {{MFC} {Datasets}: {Large}-{Scale} {Benchmark} {Datasets} for {Media} {Forensic} {Challenge} {Evaluation}},
	shorttitle = {{MFC} {Datasets}},
	url = {https://www.nist.gov/publications/mfc-datasets-large-scale-benchmark-datasets-media-forensic-challenge-evaluation},
	language = {en},
	urldate = {2026-02-26},
	journal = {NIST},
	publisher = {Haiying Guan, Mark Kozak, Eric Robertson, Yooyoung Lee, Amy Yates, Andrew Delgado, Daniel F. Zhou, Timothée N. Kheyrkhah, Jeff Smith, Jonathan G. Fiscus},
	author = {Guan, Haiying and Kozak, Mark and Robertson, Eric and Lee, Yooyoung and Yates, Amy and Delgado, Andrew and Zhou, Daniel F. and Kheyrkhah, Timothée N. and Smith, Jeff and Fiscus, Jonathan G.},
	month = jan,
	year = {2019},
	note = {Last Modified: 2019-12-31T06:12-05:00},
}

@inproceedings{dong_casia_2013,
	title = {{CASIA} {Image} {Tampering} {Detection} {Evaluation} {Database}},
	url = {https://ieeexplore.ieee.org/document/6625374},
	doi = {10.1109/ChinaSIP.2013.6625374},
	urldate = {2026-02-26},
	booktitle = {2013 {IEEE} {China} {Summit} and {International} {Conference} on {Signal} and {Information} {Processing}},
	author = {Dong, Jing and Wang, Wei and Tan, Tieniu},
	month = jul,
	year = {2013},
	pages = {422--426},
}

@inproceedings{sidere_dataset_2017,
	title = {A dataset for forgery detection and spotting in document images},
	issn = {2472-7601},
	url = {https://ieeexplore.ieee.org/document/8090394},
	doi = {10.1109/EST.2017.8090394},
	urldate = {2026-02-26},
	booktitle = {2017 {Seventh} {International} {Conference} on {Emerging} {Security} {Technologies} ({EST})},
	author = {Sidere, Nicolas and Cruz, Francisco and Coustaty, Mickal and Ogier, Jean-Marc},
	month = sep,
	year = {2017},
	note = {ISSN: 2472-7601},
	pages = {26--31},
}

@inproceedings{tornes_receipt_2023,
	address = {Berlin, Heidelberg},
	title = {Receipt {Dataset} for {Document} {Forgery} {Detection}},
	isbn = {978-3-031-41681-1},
	url = {https://doi.org/10.1007/978-3-031-41682-8_28},
	doi = {10.1007/978-3-031-41682-8_28},
	urldate = {2026-02-26},
	booktitle = {Document {Analysis} and {Recognition} - {ICDAR} 2023: 17th {International} {Conference}, {San} {José}, {CA}, {USA}, {August} 21–26, 2023, {Proceedings}, {Part} {III}},
	publisher = {Springer-Verlag},
	author = {Tornés, Beatriz Martínez and Taburet, Théo and Boros, Emanuela and Rouis, Kais and Doucet, Antoine and Gomez-Krämer, Petra and Sidere, Nicolas and d’Andecy, Vincent Poulain},
	month = aug,
	year = {2023},
	pages = {454--469},
}

@misc{dolhansky_deepfake_2020,
	title = {The {DeepFake} {Detection} {Challenge} ({DFDC}) {Dataset}},
	url = {http://arxiv.org/abs/2006.07397},
	doi = {10.48550/arXiv.2006.07397},
	urldate = {2026-02-26},
	publisher = {arXiv},
	author = {Dolhansky, Brian and Bitton, Joanna and Pflaum, Ben and Lu, Jikuo and Howes, Russ and Wang, Menglin and Ferrer, Cristian Canton},
	month = oct,
	year = {2020},
	note = {arXiv:2006.07397 [cs]},
}

@misc{he_forgerynet_2021,
	title = {{ForgeryNet}: {A} {Versatile} {Benchmark} for {Comprehensive} {Forgery} {Analysis}},
	shorttitle = {{ForgeryNet}},
	url = {http://arxiv.org/abs/2103.05630},
	doi = {10.48550/arXiv.2103.05630},
	urldate = {2026-02-26},
	publisher = {arXiv},
	author = {He, Yinan and Gan, Bei and Chen, Siyu and Zhou, Yichun and Yin, Guojun and Song, Luchuan and Sheng, Lu and Shao, Jing and Liu, Ziwei},
	month = jul,
	year = {2021},
	note = {arXiv:2103.05630 [cs]},
}

@inproceedings{le_openforensics_2021,
	title = {{OpenForensics}: {Large}-{Scale} {Challenging} {Dataset} for {Multi}-{Face} {Forgery} {Detection} and {Segmentation} {In}-the-{Wild}},
	shorttitle = {{OpenForensics}},
	url = {https://openaccess.thecvf.com/content/ICCV2021/html/Le_OpenForensics_Large-Scale_Challenging_Dataset_for_Multi-Face_Forgery_Detection_and_Segmentation_ICCV_2021_paper.html},
	language = {en},
	urldate = {2026-02-26},
	author = {Le, Trung-Nghia and Nguyen, Huy H. and Yamagishi, Junichi and Echizen, Isao},
	year = {2021},
	pages = {10117--10127},
}

@misc{liCelebDFLargescaleChallenging2020,
	title = {Celeb-{DF}: {A} {Large}-scale {Challenging} {Dataset} for {DeepFake} {Forensics}},
	shorttitle = {Celeb-{DF}},
	url = {http://arxiv.org/abs/1909.12962},
	doi = {10.48550/arXiv.1909.12962},
	urldate = {2026-02-20},
	publisher = {arXiv},
	author = {Li, Yuezun and Yang, Xin and Sun, Pu and Qi, Honggang and Lyu, Siwei},
	month = mar,
	year = {2020},
	note = {arXiv:1909.12962 [cs]},
}

@misc{corviDetectionSyntheticImages2022,
	title = {On the detection of synthetic images generated by diffusion models},
	url = {http://arxiv.org/abs/2211.00680},
	doi = {10.48550/arXiv.2211.00680},
	urldate = {2026-02-20},
	publisher = {arXiv},
	author = {Corvi, Riccardo and Cozzolino, Davide and Zingarini, Giada and Poggi, Giovanni and Nagano, Koki and Verdoliva, Luisa},
	month = nov,
	year = {2022},
	note = {arXiv:2211.00680 [cs]},
}

@misc{chesneyDeepFakesLooming2018,
	address = {Rochester, NY},
	type = {{SSRN} {Scholarly} {Paper}},
	title = {Deep {Fakes}: {A} {Looming} {Challenge} for {Privacy}, {Democracy}, and {National} {Security}},
	shorttitle = {Deep {Fakes}},
	url = {https://papers.ssrn.com/abstract=3213954},
	doi = {10.2139/ssrn.3213954},
	language = {en},
	urldate = {2026-02-20},
	publisher = {Social Science Research Network},
	author = {Chesney, Robert and Citron, Danielle Keats},
	month = jul,
	year = {2018},
}

@misc{wangCNNgeneratedImagesAre2020,
	title = {{CNN}-generated images are surprisingly easy to spot... for now},
	url = {http://arxiv.org/abs/1912.11035},
	doi = {10.48550/arXiv.1912.11035},
	urldate = {2026-02-20},
	publisher = {arXiv},
	author = {Wang, Sheng-Yu and Wang, Oliver and Zhang, Richard and Owens, Andrew and Efros, Alexei A.},
	month = apr,
	year = {2020},
	note = {arXiv:1912.11035 [cs]},
}

@misc{rosslerFaceForensicsLearningDetect2019,
	title = {{FaceForensics}++: {Learning} to {Detect} {Manipulated} {Facial} {Images}},
	shorttitle = {{FaceForensics}++},
	url = {http://arxiv.org/abs/1901.08971},
	doi = {10.48550/arXiv.1901.08971},
	urldate = {2026-02-20},
	publisher = {arXiv},
	author = {Rössler, Andreas and Cozzolino, Davide and Verdoliva, Luisa and Riess, Christian and Thies, Justus and Nießner, Matthias},
	month = aug,
	year = {2019},
	note = {arXiv:1901.08971 [cs]},
}

@article{dalalDeepfakesCourtHow2025,
	title = {Deepfakes in {Court}: {How} {Judges} {Can} {Proactively} {Manage} {Alleged} {AI}-{Generated} {Material} in {National} {Security} {Cases}},
	volume = {2024},
	issn = {0892-5593},
	shorttitle = {Deepfakes in {Court}},
	url = {https://chicagounbound.uchicago.edu/uclf/vol2024/iss1/3},
	number = {1},
	journal = {University of Chicago Legal Forum},
	author = {Dalal, Abhishek and Gao, Chongyang and Grimm, Hon Paul and Grossman, Maura and Jr, Daniel Linna and Pulice, Chiara and Subrahmanian, V. S. and Tunheim, Hon John},
	month = jan,
	year = {2025},
}

@article{delfinoDeepfakesTrialCall2023,
	title = {Deepfakes on {Trial}: {A} {Call} {To} {Expand} the {Trial} {Judge}’s {Gatekeeping} {Role} {To} {Protect} {Legal} {Proceedings} from {Technological} {Fakery}},
	volume = {74},
	issn = {0017-8322{\textless}br /{\textgreater}© Copyright University of California, College of the Law San Francisco},
	shorttitle = {Deepfakes on {Trial}},
	url = {https://repository.uclawsf.edu/hastings_law_journal/vol74/iss2/3},
	number = {2},
	journal = {UC Law Journal},
	author = {Delfino, Rebecca},
	month = feb,
	year = {2023},
	pages = {293},
}

@misc{chandraDeepfakeEval2024MultiModalIntheWild2025,
	title = {Deepfake-{Eval}-2024: {A} {Multi}-{Modal} {In}-the-{Wild} {Benchmark} of {Deepfakes} {Circulated} in 2024},
	shorttitle = {Deepfake-{Eval}-2024},
	url = {http://arxiv.org/abs/2503.02857},
	doi = {10.48550/arXiv.2503.02857},
	urldate = {2026-02-20},
	publisher = {arXiv},
	author = {Chandra, Nuria Alina and Murtfeldt, Ryan and Qiu, Lin and Karmakar, Arnab and Lee, Hannah and Tanumihardja, Emmanuel and Farhat, Kevin and Caffee, Ben and Paik, Sejin and Lee, Changyeon and Choi, Jongwook and Kim, Aerin and Etzioni, Oren},
	month = mar,
	year = {2025},
	note = {arXiv:2503.02857 [cs]
version: 1},
}

@article{bray_testing_2023,
	title = {Testing human ability to detect ‘deepfake’ images of human faces},
	volume = {9},
	issn = {2057-2085},
	url = {https://doi.org/10.1093/cybsec/tyad011},
	doi = {10.1093/cybsec/tyad011},
	number = {1},
	urldate = {2026-02-13},
	journal = {Journal of Cybersecurity},
	author = {Bray, Sergi D and Johnson, Shane D and Kleinberg, Bennett},
	month = jan,
	year = {2023},
	pages = {tyad011},
}

@misc{roca_how_2025,
	title = {How good are humans at detecting {AI}-generated images? {Learnings} from an experiment},
	shorttitle = {How good are humans at detecting {AI}-generated images?},
	url = {http://arxiv.org/abs/2507.18640},
	doi = {10.48550/arXiv.2507.18640},
	urldate = {2026-02-13},
	publisher = {arXiv},
	author = {Roca, Thomas and Roman, Anthony Cintron and Vega, Jehú Torres and Duarte, Marcelo and Wang, Pengce and White, Kevin and Misra, Amit and Ferres, Juan Lavista},
	month = may,
	year = {2025},
	note = {arXiv:2507.18640 [cs]
version: 1},
}
\appendix

\section{Generator Configuration and Pinned Versions}\label{app:tools}

The generation pipeline reads its configuration from a single \texttt{tools.yaml} file released alongside the corpus. Every manifest entry records the \texttt{tool\_specific} string of the variant that produced it, so any future drift in the underlying APIs is recoverable from the corpus release. Table~\ref{tab:pinned-versions} reports the exact pinned model identifiers for each variant; the full configuration is shipped with the v0.1.0 corpus release.
\begin{table}[h]
\centering
\small
\caption{Pinned model identifiers per variant. Variants A--C are commercial APIs whose underlying models change without notice; the \texttt{tool\_specific} string is recorded per item in the manifest. Variant D is an open-weights diffusion stack served via a local ComfyUI instance.}
\label{tab:pinned-versions}
\begin{tabular}{lll}
\toprule
Variant & Family & Model identifier \\
\midrule
A & GPT      & \texttt{openai:gpt-image-2+gpt-4o-2024-11-20} \\
B & Gemini   & \texttt{google:gemini-2.5-flash-image+gemini-2.5-pro} \\
C & Ideogram & \texttt{ideogram:v3} \\
D & Open     & \texttt{comfyui:flux.1-kontext+flux.1-dev} \\
\bottomrule
\end{tabular}
\end{table}

Variant C (Ideogram) does not expose a text-only model, so Tier 3 and Tier 4 text drafting under variant C is delegated to variant A's text endpoint via a \texttt{text\_fallback} mechanism declared in \texttt{tools.yaml}. Variant D uses the same fallback for Tier 3 language drafting; image generation under D is performed locally by the FLUX.1-Kontext-dev checkpoint (Tiers 1--3 inpainting) and the FLUX.1-dev checkpoint (Tier 4 whole-image fabrication), both retrieved from Hugging Face under their respective gated-model licenses.

The provenance configuration is also defined in \texttt{tools.yaml}: the EXIF comment string \texttt{synthetic-evidence-corpus:true}, an XMP namespace \url{https://synthetic-evidence-corpus.example/ns/1.0}, and a 64-bit LSB steganographic marker with magic bytes \texttt{5345432d53594e5448} (``SEC-SYNTH'') embedded in the first channel of every manipulated image. At least one of these three markers is designed to survive a re-save through a consumer image editor.

\section{Prompt Strategies by Tier}\label{app:prompts}

This appendix describes the prompt strategies used at each tier. The verbatim prompt templates are recorded in the per-item manifest and included in the corpus release; what follows is the structural description that lets a reader understand how each template is parameterised.

\paragraph{Tier 1 (single-field edits).} Most Tier 1 manipulations are mechanical and do not invoke an LLM: date and dollar-amount changes are applied by deterministic local operations on the document image or text. A prompt is invoked only for the variant-A/B/C image-edit path, which performs full-frame inpainting on the receipt or document image with a clone-style instruction parameterised by the source field value, the target field value, and the bounding box of the edit region. The Tier 1 dollar sub-variant produces both \emph{consistent} (subtotals/taxes/totals recomputed) and \emph{inconsistent} (only the line item and total updated) versions in equal proportion, controlled by a sub-variant flag rather than prompt content.

\paragraph{Tier 2 (forged handwriting).} Tier 2 prompts request the inpainting of a signature or short handwritten annotation in a specified region. The prompt parameters are the chosen style (drawn from a 200-style pool partitioned 150 train / 50 test), the rendered glyph image of the target name (drawn fresh per item via Faker), and the region's bounding box. The style image and the rendered glyphs serve as visual references for the variant-A/B/C image generators and the variant-D Flux.1-Kontext workflow alike. Names are independent of style indices, so name identity does not leak style identity.

\paragraph{Tier 3 (inserted content).} Tier 3 prompts have two stages. A first stage drafts the inserted content (a contract clause, a memo or report paragraph, or a new reply in an existing email thread) using a fixed prompt template parameterised by the substrate document family, the surrounding context extracted from the source artifact, and the intended evidentiary role. A second stage inserts the drafted text at a plausible location in the source artifact, with font and indentation matched to the surrounding text, via the variant's inpainting path. The email-insertion case warrants particular mention: the inserted reply impersonates a participant already present in the thread (the prompt is parameterised by the participant's prior messages and signature style) rather than introducing a fresh fictional identity, since dropping a stranger into a real thread produces an obviously wrong training signal.

\paragraph{Tier 4 (whole fabrication).} Tier 4 prompts request a complete document of a specified family with no source artifact. Three to five documents from the corresponding source dataset are passed to the generator as style anchors (few-shot image conditioning), so that the output belongs plausibly to the target genre. The prompt is parameterised by the target family, the fabricated identity bundle (names, employers, addresses, email addresses, phone numbers, and, for medical-style documents, NPIs with explicitly invalid Luhn checksums) supplied per item via the Faker-based identity library, and, for fabricated business documents, the parametrically generated letterhead. For fabricated receipts the prompt produces arithmetically consistent and arithmetically broken sub-variants in equal proportion, again controlled by a sub-variant flag. 

\section{Software Dependencies}\label{app:dependencies}

The generation pipeline is implemented in Python 3.11 and orchestrated by the \texttt{sec} package (released alongside the corpus). The external dependencies on which the pipeline relies are listed in Table~\ref{tab:dependencies}. The pipeline is structured as a sequence of phases (infrastructure setup, training-pool generation, test-pool generation, manifest validation, provenance audit) coordinated by per-source orchestration scripts (one each for SROIE, CORD, and FindIt2) that share a common batch-running and adapter-loading layer.

\begin{table}[h]
\centering
\small
\caption{External software dependencies of the generation pipeline. Versions are pinned in the released code's \texttt{requirements.txt}; pinned ranges are reported here.}
\label{tab:dependencies}
\begin{tabular}{lll}
\toprule
Package & Purpose & Version \\
\midrule
Python                  & Runtime                                   & 3.11 \\
PyArrow                 & Manifest read/write (Parquet)             & $\geq$ 14.0 \\
PyYAML                  & Configuration loading (\texttt{tools.yaml}, sidecars) & $\geq$ 6.0 \\
Pillow                  & Image manipulation, EXIF/XMP handling     & $\geq$ 10.0 \\
Faker                   & Per-item identity generation              & $\geq$ 24.0 \\
Requests                & HTTP client (Ideogram, ComfyUI)           & $\geq$ 2.31 \\
Tesseract               & OCR for FindIt2 substrate parsing         & $\geq$ 5.0 \\
\texttt{huggingface\_hub} & Variant D checkpoint download           & $\geq$ 0.20 \\
ComfyUI                 & Variant D workflow execution               & latest from upstream \\
FLUX.1-Kontext-dev         & Variant D inpainting checkpoint           & via Hugging Face \\
FLUX.1-dev              & Variant D whole-image checkpoint          & via Hugging Face \\
OpenAI Python SDK       & Variant A API client                       & $\geq$ 1.0 \\
Google GenerativeAI SDK & Variant B API client                       & $\geq$ 0.5 \\
\bottomrule
\end{tabular}
\end{table}

The variant D checkpoints (FLUX.1-Kontext-dev and FLUX.1-dev) are gated on Hugging Face and require license acceptance through the Hugging Face account before download; a helper script (\texttt{download\_comfyui\_checkpoints.sh}) is provided in the
release. ComfyUI workflows for the two checkpoints are released as JSON files (\texttt{flux\_kontext\_edit.json} and
\texttt{flux1\_dev\_style\_generate.json}) referenced from \texttt{tools.yaml}. The corpus's deterministic seeding scheme uses
two base seeds defined in \texttt{tools.yaml}: \texttt{base\_train = 11000} for the training pool and \texttt{base\_test = 81000} for the test pool, with per-item seeds derived as $\text{batch\_seed} \times 1000 + \text{item\_index}$ as described in Section~\ref{sec:generation}.

\section*{NeurIPS Paper Checklist}

\begin{enumerate}

\item {\bf Claims}
    \item[] Question: Do the main claims made in the abstract and introduction accurately reflect the paper's contributions and scope?
    \item[] Answer: \answerYes{}
    \item[] Justification: The abstract and introduction describe the corpus's scope (three document families, four manipulation tiers, four generator families, 2{,}490 items), its three design commitments (source-disjoint pools, ablation-ready shortcut controls, provenance markers), and explicitly disclaim what the corpus does not establish. Limitations are stated in Section~\ref{sec:limitations}.

\item {\bf Limitations}
    \item[] Question: Does the paper discuss the limitations of the work performed by the authors?
    \item[] Answer: \answerYes{}
    \item[] Justification: Limitations are discussed in a dedicated Section~\ref{sec:limitations}, covering scope (three families, single manipulation per item), the gap between synthetic and real-world fraud, tool-version drift in commercial APIs, the lab-to-wild performance gap, and the corpus's inability to support cross-document consistency analysis.

\item {\bf Theory assumptions and proofs}
    \item[] Question: For each theoretical result, does the paper provide the full set of assumptions and a complete (and correct) proof?
    \item[] Answer: \answerNA{}
    \item[] Justification: The paper introduces a dataset and does not present theoretical results.

    \item {\bf Experimental result reproducibility}
    \item[] Question: Does the paper fully disclose all the information needed to reproduce the main experimental results of the paper to the extent that it affects the main claims and/or conclusions of the paper (regardless of whether the code and data are provided or not)?
    \item[] Answer: \answerYes{}
    \item[] Justification: The paper is a dataset contribution. Section~\ref{sec:dataset} fully documents the construction pipeline including substrate sources (Table~\ref{tab:sources}), per-tier manipulation procedures, generator versions pinned in \texttt{tools.yaml}, deterministic per-item seed derivation, and the manifest schema. Variant D (Flux.1-Kontext-dev / Flux.1-dev) is fully reproducible without commercial-provider cooperation; Variants A--C are reproducible only insofar as the commercial APIs remain stable, a limitation acknowledged in Section~\ref{sec:limitations}.

\item {\bf Open access to data and code}
    \item[] Question: Does the paper provide open access to the data and code, with sufficient instructions to faithfully reproduce the main experimental results, as described in supplemental material?
    \item[] Answer: \answerYes{}
    \item[] Justification: The corpus is hosted on Harvard Dataverse at \url{https://doi.org/10.7910/DVN/YY0IUH} and will be publicly accessible by the camera-ready deadline. A Croissant metadata file accompanying the release documents the manifest schema, generator versions, and per-item provenance for downstream use.

\item {\bf Experimental setting/details}
    \item[] Question: Does the paper specify all the training and test details (e.g., data splits, hyperparameters, how they were chosen, type of optimizer) necessary to understand the results?
    \item[] Answer: \answerNA{}
    \item[] Justification: The paper introduces a dataset and does not report detector training experiments. Baseline detector evaluation is identified as future work (Section~\ref{sec:future_work}).

\item {\bf Experiment statistical significance}
    \item[] Question: Does the paper report error bars suitably and correctly defined or other appropriate information about the statistical significance of the experiments?
    \item[] Answer: \answerNA{}
    \item[] Justification: The paper does not report detector training or evaluation experiments.

\item {\bf Experiments compute resources}
    \item[] Question: For each experiment, does the paper provide sufficient information on the computer resources (type of compute workers, memory, time of execution) needed to reproduce the experiments?
    \item[] Answer: \answerNA{}
    \item[] Justification: The paper introduces a dataset and does not report training or evaluation experiments.
    
\item {\bf Code of ethics}
    \item[] Question: Does the research conducted in the paper conform, in every respect, with the NeurIPS Code of Ethics \url{https://neurips.cc/public/EthicsGuidelines}?
    \item[] Answer: \answerYes{}
    \item[] Justification: The work conforms to the NeurIPS Code of Ethics. Substrate corpora are used under their original licenses. The corpus carries provenance markers on every artifact to prevent misuse as genuine evidence. PII inherited from substrates is acknowledged in the dataset's Croissant metadata. Ethical considerations specific to a synthetic-evidence corpus are discussed in Section~\ref{sec:discussion}.

\item {\bf Broader impacts}
    \item[] Question: Does the paper discuss both potential positive societal impacts and negative societal impacts of the work performed?
    \item[] Answer: \answerYes{}
    \item[] Justification: Both positive and negative impacts are addressed throughout. The corpus is motivated by the access-to-justice gap created by AI-manipulated evidence (Sections~\ref{sec:Introduction} and~\ref{sec:Background}) and aims to support detection infrastructure for self-represented and under-resourced litigants. Negative-impact risks are addressed by redundant provenance markers (Section~\ref{sec:generation}) ensuring no corpus artifact can be mistaken for genuine evidence, and by source-level safeguards on PII inherited from substrates.
    
\item {\bf Safeguards}
    \item[] Question: Does the paper describe safeguards that have been put in place for responsible release of data or models that have a high risk for misuse (e.g., pre-trained language models, image generators, or scraped datasets)?
    \item[] Answer: \answerYes{}
    \item[] Justification: Section~\ref{sec:generation} describes a three-redundant provenance marking scheme (EXIF/XMP tag, steganographic flag, sentinel hash) applied to every artifact, designed so that at least one marker survives a re-save through a consumer image editor. Fictional identities used throughout the corpus are generated deterministically via Faker rather than drawn from real persons.

\item {\bf Licenses for existing assets}
    \item[] Question: Are the creators or original owners of assets (e.g., code, data, models), used in the paper, properly credited and are the license and terms of use explicitly mentioned and properly respected?
    \item[] Answer: \answerYes{}
    \item[] Justification: Substrate corpora are credited throughout Section~\ref{sec:collection} with citations to original publications. Source datasets used: SROIE, CORD-v2, Find It Again!, Enron, RVL-CDIP, DUDE, Avocado (LDC2015T03), and the UCSF Industry Documents Library. Each source's license is recorded per item in the manifest, and pool assignment respects each source's redistribution terms.

\item {\bf New assets}
    \item[] Question: Are new assets introduced in the paper well documented and is the documentation provided alongside the assets?
    \item[] Answer: \answerYes{}
    \item[] Justification: The CIFAR Synthetic Evidence Corpus is the new asset introduced. Documentation includes the paper itself, the Croissant metadata file accompanying the Dataverse release, the per-item Parquet manifest documenting generator versions, prompts, edit regions, and per-item seeds, and the \texttt{tools.yaml} file pinning exact generator versions.

\item {\bf Crowdsourcing and research with human subjects}
    \item[] Question: For crowdsourcing experiments and research with human subjects, does the paper include the full text of instructions given to participants and screenshots, if applicable, as well as details about compensation (if any)? 
    \item[] Answer: \answerNA{}
    \item[] Justification: The work does not involve crowdsourcing or research with human subjects. All manipulated artifacts are produced through the automated generation pipeline described in Section~\ref{sec:dataset}.

\item {\bf Institutional review board (IRB) approvals or equivalent for research with human subjects}
    \item[] Question: Does the paper describe potential risks incurred by study participants, whether such risks were disclosed to the subjects, and whether Institutional Review Board (IRB) approvals (or an equivalent approval/review based on the requirements of your country or institution) were obtained?
    \item[] Answer: \answerNA{}
    \item[] Justification: The work does not involve human subjects research.

\item {\bf Declaration of LLM usage}
    \item[] Question: Does the paper describe the usage of LLMs if it is an important, original, or non-standard component of the core methods in this research? Note that if the LLM is used only for writing, editing, or formatting purposes and does \emph{not} impact the core methodology, scientific rigor, or originality of the research, declaration is not required.
    \item[] Answer: \answerYes{}
    \item[] Justification: Large language models are a core component of the dataset construction pipeline. Section~\ref{sec:generation} documents their use: GPT-4o, Gemini 2.5 Pro, and Ideogram 3.0 are used as generator variants A, B, and C respectively to produce manipulated artifacts. LLMs are also used to draft the inserted content for Tier 3. Exact model versions are pinned in \texttt{tools.yaml} and recorded per item in the manifest.

\end{enumerate}

\end{document}